\documentclass[a4paper,french]{rnti}

\usepackage[T1]{fontenc}
\usepackage[latin1,utf8]{inputenc}

\usepackage{url}

\usepackage{graphicx}

\nomcourt{G. Poux-Médard et al.}

\titre{Le Processus Powered Dirichlet-Hawkes comme A Priori Flexible pour Clustering Temporel de Textes}
\titrecourt{Powered Dirichlet-Hawkes Process}

\auteur{Gaël Poux-Médard\affil{1},
        Julien Velcin\affil{1},
        Sabine Loudcher\affil{1}}

\affiliation{
    \affil{1}Université de Lyon, Lyon 2, ERIC UR 3083\\
    5 avenue Pierre Mendès France, F69676 Bron Cedex, France\\
    \http{http://eric.univ-lyon2.fr}
 }

\resume{%
Le contenu textuel d’un document et sa date de publication sont corrélés. Par exemple, une publication scientifique est influencée par les précédents articles cités dans ladite publication. Utiliser cette corrélation permet d’améliorer la compréhension de grands corpus textuel datés. Cependant, cette tâche peut se compliquer lorsque les textes considérés sont courts ou possèdent des vocabulaires similaires. De plus, la corrélation entre texte et date est rarement parfaite.
Nous développons une méthode répondant à ces limites, permettant de créer des clusters de documents en fonction de leur contenu et de leur date : le processus Powered Dirichlet-Hawkes (PDHP). Nous montrons que PDHP présente de meilleures performances que les modèles état de l’art (qu’il généralise) lorsque l’information textuelle ou temporelle est peu informative. Le PDHP se libère également de l’hypothèse d’une corrélation parfaite entre texte et date des documents. Enfin, nous illustrons une possible application sur des données réelles, provenant de Reddit.
}

\summary{%
The textual content of a document and its publication date are intertwined. For example, the publication of a news article on a topic is influenced by previous publications on similar issues, according to underlying temporal dynamics. However, it can be challenging to retrieve meaningful information when textual information conveys little. Furthermore, the textual content of a document is not always correlated to its temporal dynamics.
We develop a method to create clusters of textual documents according to both their content and publication time, the Powered Dirichlet-Hawkes process (PDHP). PDHP yields significantly better results than state-of-the-art models when temporal information or textual content is weakly informative. PDHP also alleviates the hypothesis that textual content and temporal dynamics are perfectly correlated. We demonstrate that PDHP generalizes previous work --such as DHP and UP. Finally, we illustrate a possible application using a real-world dataset from Reddit.
}

\usepackage{amsmath,amssymb,amsfonts}
\usepackage{algorithmic}
\usepackage{graphicx}
\usepackage{textcomp}
\usepackage{xcolor}

\newcommand{\donotdisplay}[1]{}
\graphicspath{{Figures/}}
\usepackage{cancel}

\begin{document}

%
\section{Introduction}
Les contenus numériques sont générés à une vitesse sans précédent. Chaque minute, environ 500~000 commentaires sont postés sur Facebook, 400h de vidéo sont mises en ligne sur Youtube, et 500~000 messages sont publiés sur Twitter. Une approche possible pour comprendre cette masse d'informations est de regrouper ces publication en groupes (clusters) thématiques. Grouper des publications similaires aiderait par exemple à automatiser la détection non-supervisée de thématiques ou à générer des résumés de nouvelles journalières. 

De récents travaux ont montré que considérer la date de publication augmente les performances des algorithmes de clustering \citep{Du2012KernelCascade}.
La plupart de ces modèles fonctionnent par échantillonnage : une observation récente aura plus de chances d'être utilisée dans l'apprentissage qu'une observation éloignée dans le temps \citep{Amr2008RCRP,Blei2006DynamicTopicModel,Yin2018ShortTextDHP}. Cependant, cela implique que la fonction d'échantillonnage temporel transcrive bien la réalité des dynamiques à l'oeuvre, ce qui n'est pas évident. En outre, ces modèles se basent sur un \textit{a priori} de Dirichlet (DP) pour créer les clusters. Hors, il a été montré que DP n'est pas assez flexible pour décrire des processus en temps continu. Dans \citep{Du2015DHP}, les auteurs dérivent le processus de Dirichlet-Hawkes (DHP), et l'utilisent comme \textit{a priori} bayésien pour grouper des documents en considérant leur date de publication en temps continu. Cependant, de récentes études montrent que cette méthode trouve ses limites dans les cas où le contenu textuel est peu informatif (textes courts, vocabulaires similaires entre thématiques, etc.) \citep{Yin2018ShortTextDHP}. 

Notre travail vise donc à dépasser ces limites en développant le processus Powered Dirichlet Hawkes (ou Powered Dirichlet-Hawkes process, PDHP). Nous montrons qu'il existe d'autres contextes dans lesquels DHP échoue à rendre compte du processus à l'oeuvre, là où PDHP présente de bonnes performances. Typiquement, DHP échoue dans les cas où les dates de publication sont peu informatives.
Nous montrons également qu'il existe des cas où l'information textuelle est décorrélée des dynamiques de publication, ce que DHP ne considère pas. Par exemple, un article publié par un journal influent aura bien plus de résonance qu'un article strictement identique publié par un journal moins populaire ; l'influence de cet article ne dépend pas uniquement de son contenu, mais également de sa date. Nous répondons à toutes ces limitations avec le PDHP et démontrons sont efficacité sur plusieurs jeux de données (jusqu'à +0.3 NMI par rapport à DHP). Notre méthode permet également de distinguer les clusters \textit{textuels} des clusters \textit{temporels} lorsque le texte et la dynamique ne sont pas parfaitement corrélés.

Ce travail s'articule comme suit : nous brossons d'abord un court état de l'art des méthodes de clustering temporel et des alternatives aux processus de Dirichlet. Nous formulons ensuite notre \textit{a priori} mathématiquement, puis l'associons à un modèle de langue simple. Enfin, nous effectuons des expériences sur des jeux de données synthétiques et réels, et résumons les résultats obtenus, avant de conclure en ouvrant sur de possibles extensions du PDHP.
Explicitement, listons nos contributions :
\begin{itemize}
    \item Explication des limites de DHP : quand le contenu textuel ou temporel est peu informatif. En outre le modèle suppose une corrélation parfaite entre ces deux variables.
    \item Dérivation du processus Powered Dirichlet-Hawkes (PDHP), qui généralise DHP et le processus Uniforme (UP).
    \item Démonstration par une étude systématique sur plusieurs jeux de données que PDHP fournit de meilleurs résultats que DHP et UP.
    \item Démonstration que PDHP permet d'ajuster l'importance donnée à l'information textuelle ou temporelle dans de la création des clusters de documents.
\end{itemize}

\section{Contexte}
\textbf{Pourquoi considérer le temps} --- La date de publication d'un document fournit une information utile aux problèmes de clustering. Souvent, l'apparition d'un nouveau document a été conditionnée par l'existence des documents précédents. Par exemple, dans la publication scientifique : le présent article est construit grâce à l'existence des références qu'il cite, et qui traitent d'un sujet proche sinon similaire. En 2012, il a été montré que l'ordre d'apparition (ou la date) de tweets modifie notre réaction aux tweets suivants \citep{Myers2012CoC}. En outre, la dimension temporelle n'importe que pour certains groupes bien définis d'information \citep{Poux2021interactions} et ne perdure pas dans le temps \citep{Cao2019AdsDataset}. 
Il est donc nécessaire d'allier la notion de cluster à la notion de dynamique pour parvenir à une description correcte des processus de publication en ligne.

\textbf{Clustering temporel de documents textuels} --- Plusieurs modèles ont été proposés pour modéliser une évolution dynamique de clusters textuels \citep{Blei2006DynamicTopicModel,Wang2006TopicsOverTime,Amr2008RCRP,Blei2010DDCRP}. Cependant, la plupart d'entre eux fonctionnent en fixant le nombre de clusters à inférer au préalable, ou approximent les dynamiques temporelles via une fonction d'échantillonnage. ce qui ne permet pas de modéliser des dynamiques temporelles. En 2015, N. Du \textit{\& al} ont proposé le DHP pour considérer des comptes non-entiers dans un espace de temps continu (en se passant d'échantillonnage temporel). L'idée clé est de remplacer les comptes du processus de Dirichlet standard par des fonctions d'intensité provenant des processus de Hawkes, dépendant du temps, résultant en un processus de Dirichlet-Hawkes (DHP).

\textbf{Alternatives aux processus de Dirichlet} --- DHP (et ses variantes) utilisent comme base un processus de Dirichlet (DP) classique. Cependant, DP est un choix arbitraire \citep{Welling2006AlterDP}. \citep{Wallach2010UnifP} a proposé le processus Uniforme (UP), puis \citep{Poux2021PDP} l'a généralisé avec le processus Powered Dirichlet (PDP) ; les variantes UP et PDP fonctionnent mieux que DP pour plusieurs jeux de données. Comme PDP généralise UP et DP, nous proposons PDHP comme généralisation de DHP et UP. Comme nous le verrons, cela nous permettra de pallier les limites de DHP lorsqu'il s'agit de textes courts \citep{Yin2018ShortTextDHP} ou de dynamiques décorrélées du contenu textuel.

\section{PDHP : Powered Dirichlet-Hawkes Process}
\subsection{PDP: Powered Dirichlet Process}
Dans un processus de Dirichlet, une nouvelle observation aura une chance \textit{a priori} d'appartenir à chaque cluster existant proportionnelle au nombre d'observations qui y appartiennent --``les riches deviennent plus riches''. Le processus plus général décrit dans \citep{Poux2021PDP} propose de modifier cette dépendance linéaire par une forme permettant une dépendance en puissance de $r$. Dans la formulation proposée, pour $r=1$ le processus est identique à un processus de Dirichlet (DP), et pour $r=0$ il est identique au processus uniforme (UP). Mathématiquement, si $C_i$ est le cluster choisi par la $i^{\text{\`eme}}$ observation, $\vec{C^-}$ l'historique d'allocation des $i-1$ observations précédentes, $N_k$ la population du cluster $k$ parmi $K$ clusters non-vides, et $\alpha_0$ un paramètre, on peut écrire le Powered Dirichlet Process (PDP) :
\begin{equation}
\label{eq-PCRP}
    \text{PDP} (C_i = c \vert r, \vec{C^-}, \alpha_0) = 
    \begin{cases}
    \frac{N_c^r}{\alpha_0 + \sum_k^K N_k^r} \text{ si c = 1, 2, ..., C}\\
    \frac{\alpha_0}{\alpha_0 + \sum_k^K N_k^r} \text{ si c = C+1}
    \end{cases}
\end{equation}

\subsection{Processus de Hawkes}
Le processus de Hawkes est une classe de processus du point (\textit{point process)} définie comme auto-stimulées ; la probabilité d'un nouvel événement dépend de la réalisation d'événements précédents. Un tel processus est entièrement caractérisé par sa fonction d'intensité $\lambda (t)$, reliée à la probabilité qu'un événement survienne entre $t$ et $t+\Delta t$ par $\lambda(t) \stackrel{\Delta t \rightarrow 0}{=} \frac{P(t_{event} \in [t;t+\Delta t])}{\Delta t}$. Dans notre modèle, chaque cluster se voit associer une telle fonction d'intensité $\lambda_c(t \vert \mathcal{H}_{<t, c})$, où $\mathcal{H}_{<t, c}$ est l'historique des événements survenus avant $t$ dans le cluster $c$. Hypothèse est faite que les événements au sein d'un même cluster provoqueront les futurs événements appartenant à ce même cluster. Explicitement :
\begin{equation}
\label{eq-HawkesClus}
    \lambda_c(t \vert \mathcal{H}_{<t, c}) = \sum_{t_{i,c} \in \mathcal{H}_{<t, c}} \vec{\alpha_c}^T \cdot \vec{\kappa}(t_{i,c})
\end{equation}
où $\vec{\alpha_c}$ est un vecteur de coefficients que l'on cherchera à inférer, et $\vec{\kappa}(t)$ un vecteur de fonctions prédéfinies (\textit{kernel functions}) dépendant du temps. 
La vraisemblance associée avec le processus de Hawkes de chaque cluster est détaillée dans la littérature associée \citep{rizoiu2017Hawkes}.
\begin{figure*}
    \centering
    \includegraphics[width=\textwidth]{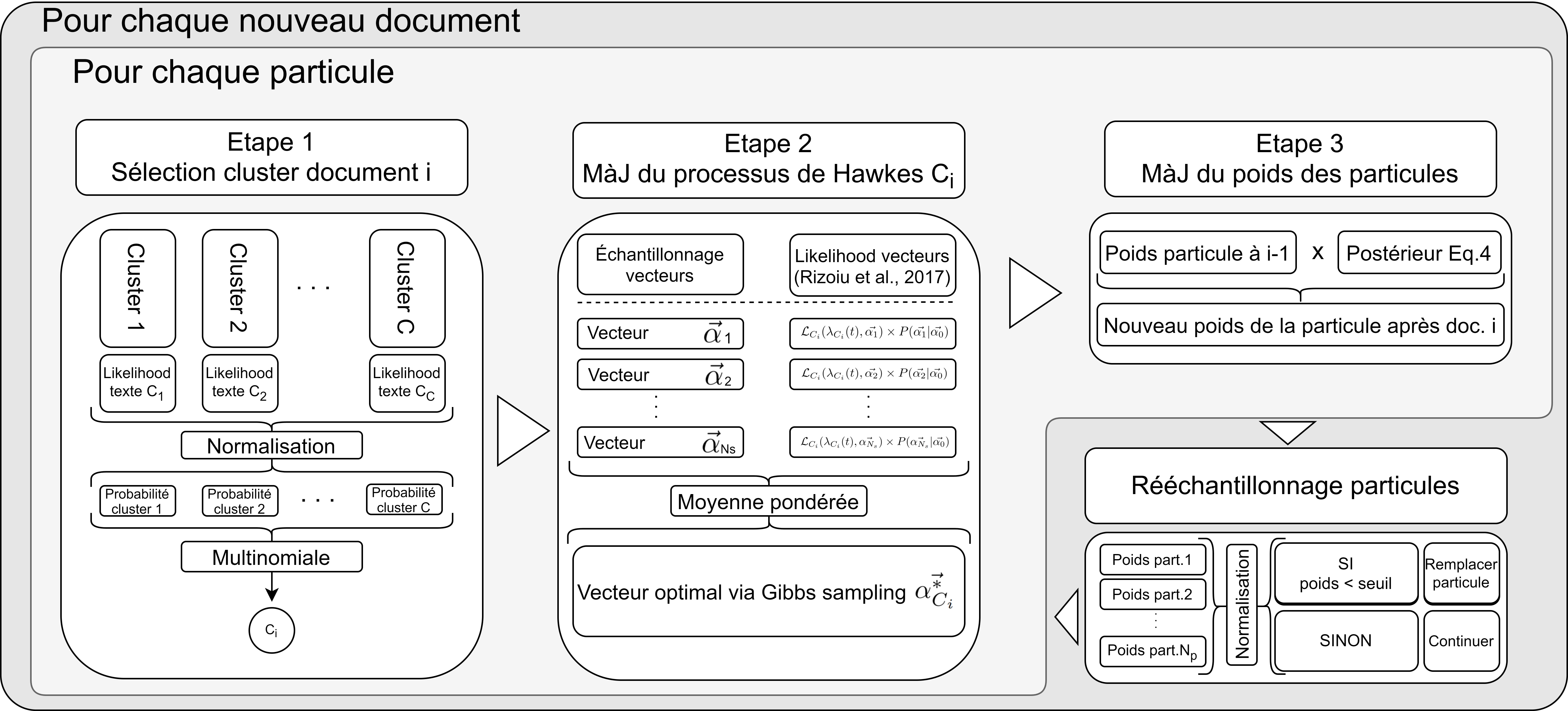}
    \caption{\textbf{Fonctionnement de l'algorithme} --- Pour chaque nouveau document, on exécute les étapes 1 (échantillonnage du cluster), 2 (mise à jour de la dynamique du cluster) et 3 (mise à jour de la plausibilité des particules) pour chaque particule, puis on remplace les particules trop peu plausibles par d'autres plus plausibles.}
    \label{fig-SchemaSMC}
\end{figure*}

\subsection{PDHP : Powered Dirichlet Hawkes Process}
Suivant le raisonnement effectué dans \citep{Du2015DHP}, nous substituons les comptes du processus Powered Dirichlet par les fonctions d'intensité des processus de Hawkes, ce qui permet d'avoir une dépendance non-linéaire ($r$) avec les fonctions d'intensité associées à chaque cluster. Nous aboutissons à la forme suivante du PDHP :
\begin{equation}
\label{eq-likModelTmp}
    P(C_i = c\vert t_i, r, \lambda_0, \mathcal{H}_{<t_i,c}) = 
    \begin{cases}
    \frac{\lambda_c^r(t_i)}{\lambda_0 + \sum_{c'} \lambda_{c'}^r(t_i)} \text{ si c$\leq$C}\\
    \frac{\lambda_0}{\lambda_0 + \sum_{c'} \lambda_{c'}^r(t_i)} \text{ si c=C+1}
    \end{cases}
\end{equation}

Nous rappelons que le PDHP a pour vertu d'être utilisé comme \textit{a priori} bayésien, et doit donc être couplé à un autre modèle --en l'occurrence, un modèle de langue. Nous choisissons de considérer un simple modèle Dirichlet-Multinomial pour modéliser le contenu textuel des documents, par soucis de simplicité. Nous pouvons écrire explicitement \textit{a posteriori} bayésien de notre modèle. Soient $N_c$ le nombre total de mots dans le cluster $c$ avant la $i^{\text{\`eme}}$ observation, $n_i$ le nombre total de mots dans le document $i$, $N_{c,v}$ de nombre d'occurrences du mot $v$ dans le cluster $c$, $n_{i,v}$ le nombre d'occurrences du mot $v$ dans le document $i$ et $\theta_0 = \sum_v \theta_{0,v}$ le paramètre de concentration du modèle Dirichlet-Multinomial. :
\begin{equation}
\label{eq-likModelTot}
\begin{split}
    P&(C_i = c \vert r, n_i, t_i, N_c, \mathcal{H}_{<t, c})
    \propto \underbrace{P(n_i \vert C_i=c, N_{<i,c}, \theta_0)}_{\text{Vraisemblance textuelle}} \underbrace{P(C_i = c\vert t_i, r, \lambda_0, \mathcal{H}_{<t_i,c})}_{\text{A priori temporel}} \\
    = &\frac{\Gamma(N_c+\theta_0)}{\Gamma(N_c+n_i+\theta_0)} \prod_v \frac{\Gamma(N_{c,v} + n_{i,v} + \theta_{0,v})}{\Gamma(N_{c,v}+\theta_0)}
    \times \begin{cases}
    \frac{\lambda_c^r(t_i)}{\lambda_0 + \sum_{c'} \lambda_c'^r(t_i)} \text{ si c = 1, ..., C}\\
    \frac{\lambda_0}{\lambda_0 + \sum_{c'} \lambda_{c'}^r(t_i)} \text{ si c = C+1}
    \end{cases}
\end{split}
\end{equation}

Abaisser la valeur de $r$ se traduit par une modification de l'importance accordée à la dynamique dans notre modélisation. Une faible valeur de $r$ lissera les différences entre les fonctions d'intensité des clusters (pour $r=0$, la valeur de l'\textit{a priori} est identique pour tous les clusters -- c'est le processus uniforme), si bien que l'allocation à un cluster se fera sur la base du contenu textuel uniquement. Au contraire, une grande valeur de $r$ exacerbe les différences d'intensité temporelle entre les clusters, et donc force l'allocation à se baser sur l'aspect temporel, alors discriminant. 
C'est ici que se trouve la principale contribution de notre méthode : régler $r$ permet de choisir si les clusters doivent être formés sur la base de l'information textuelle, temporelle, ou d'une mixture des deux.

Pour l'inférence, nous utilisons un algorithme à particules similaire à celui de \citep{Du2015DHP,Valera2017HDHP,Poux2021PDHP}. Chaque particule représente une hypothèse d'allocation de documents à des clusters. Les particules représentant des hypothèses trop peu plausibles sont remplacées par d'autres à chaque itération. Nous schématisons son exécution Fig.\ref{fig-SchemaSMC}. Plus de détails dans \cite{Poux2021PDHP} et dans l'implémentation\footnote{Données et codes disponibles sur \http{https://github.com/GaelPouxMedard/PDHP}}.
\begin{figure}
    \centering
    \includegraphics[width=0.92\columnwidth]{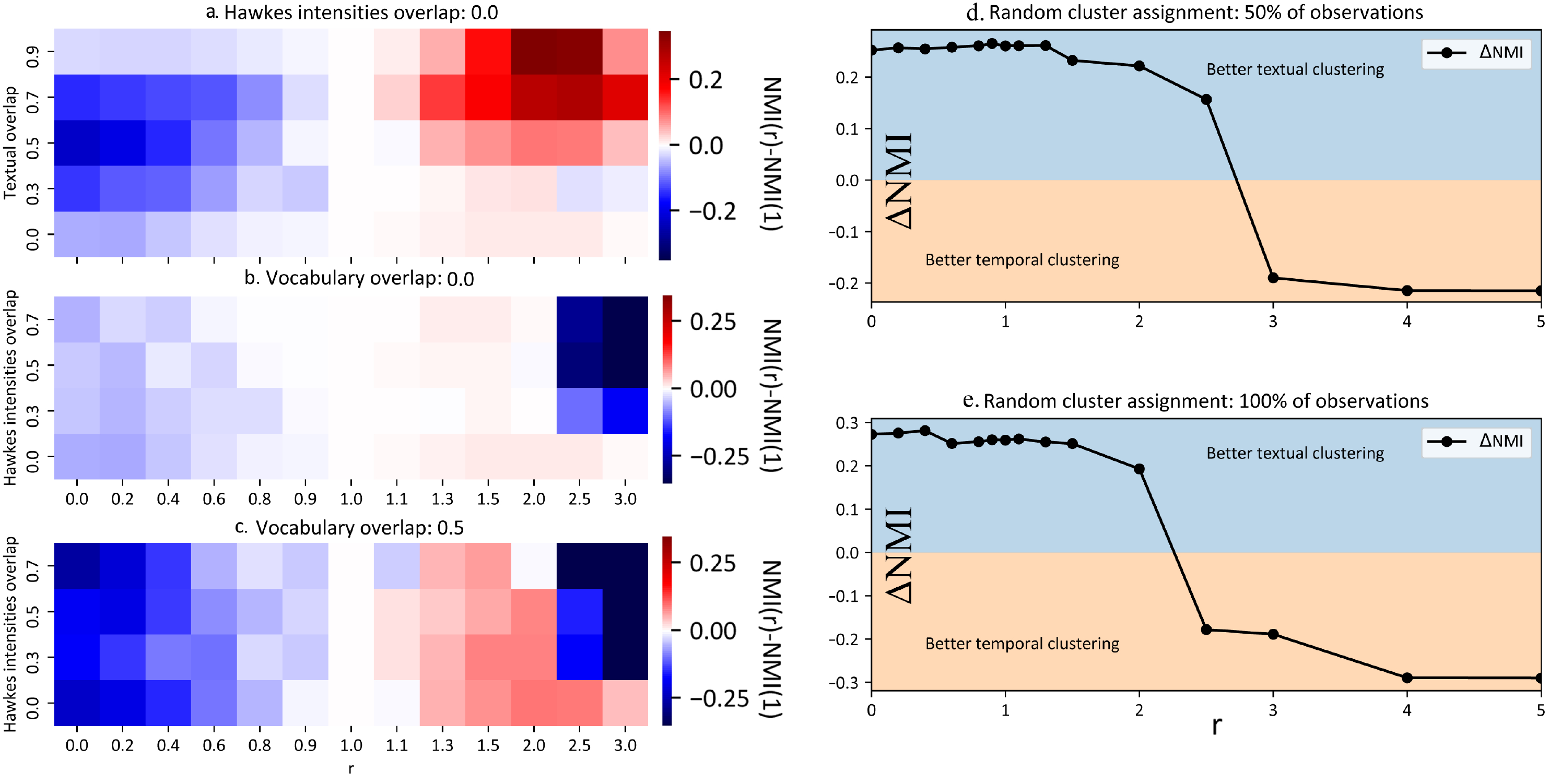}
    \caption{\textbf{PDHP permet d'améliorer les performances de l'état de l'art} --- \textbf{\textit{(Gauche)}} Différence de NMI entre PDHP et DHP \cite{Du2015DHP} pour plusieurs valeurs de $r$ et d'overlap textuel et temporel, moyenné sur 10 jeux de données par configuration. \textbf{\textit{(Droite)} Varier $r$ permet de choisir entre un clustering textuel ou temporel} --- La ligne noire représente le différence entre la NMI temporelle et la NMI textuelle. Pour de petits $r$, le clustering textuel est meilleur que le temporel ; pour de grands $r$, l'inverse.
    }
    \label{fig-res-overlaps}
\end{figure}

\section{Expériences}
\subsection{Données}
\textbf{Données synthétiques} --- Nous générons des données en utilisant deux clusters. Chaque cluster a son vocabulaire propre (1000 mots), qu'il partage dans une proportion donnée avec l'autre cluster (\textit{textual overlap}). Pour chaque cluster, nous simulons un processus de Hawkes, et associons à chaque événement 20 mots tirés parmi le vocabulaire de son cluster. Les fonctions d'intensité résultantes se chevauchent sur l'axe temporel dans une certaine proportion (``\textit{Hawkes intensity overlap}'').  Pour différentes valeurs d'\textit{overlaps}, nous générons 10 jeux de données différents. Nous évaluons nos performances en considérant la NMI (\textit{normalized mutual information}). Les autres métriques testées donnent des résultats similaires et ne sont donc pas reportées ici. En outre, nous simulons une décorrélation entre la dynamique temporelle et le contenu textuel. Pour des \textit{overlaps} nuls, nous sélectionnons au hasard un certain pourcentage d'événements et leur associons un nouveau vocabulaire tiré d'un cluster choisi aléatoirement. Nous avons alors deux types de groupes : celui qui a servi a générer le contenu textuel (cluster textuel) et celui qui a déterminé la date de publication (cluster temporel).

\textbf{Données réelles} --- Nous considérons également un jeu de données provenant de Reddit, qui fera ici office d'illustration notre méthode, constitué de 73.000 titres de posts provenant de 9 subreddits d'information en avril 2019 avec date de publication associée.

\subsection{Résultats}
Nous reportons les résultats sur les \textbf{corpus synthétiques} Fig.~\ref{fig-res-overlaps}. Nos expériences mènent aux conclusions suivantes :
\begin{itemize}
    \item PDHP permet de meilleures performances par rapport à DHP et UP lorsque l'information textuelle est faible (vocabulaires qui se recoupent, textes courts) : gain jusqu'à +0.3 de NMI par rapport à DHP (Fig.~\ref{fig-res-overlaps}a). PDHP permet une faible amélioration des résultats lorsque les données temporelles sont informatives, c'est à dire lorsque \textit{Hawkes intensity overlap} est nul (Fig.~\ref{fig-res-overlaps}b). Enfin, dans des situations plus réalistes où à la fois le vocabulaire et les intensités se recoupent, PDHP permet de meilleurs résultats : gain jusqu'à +0.2 de NMI par rapport à DHP (Fig.~\ref{fig-res-overlaps}c).
    \item Fig.~\ref{fig-res-overlaps}(d,e). Lorsque le contenu textuel est décorrelé de la dynamique de publication (ce qui se rapproche des processus de diffusion d'information réels), varier $r$ permet de modéliser avec de bonnes performances l'une ou l'autre information.
\end{itemize}

\begin{figure}
    \centering
    \includegraphics[width=0.85\columnwidth]{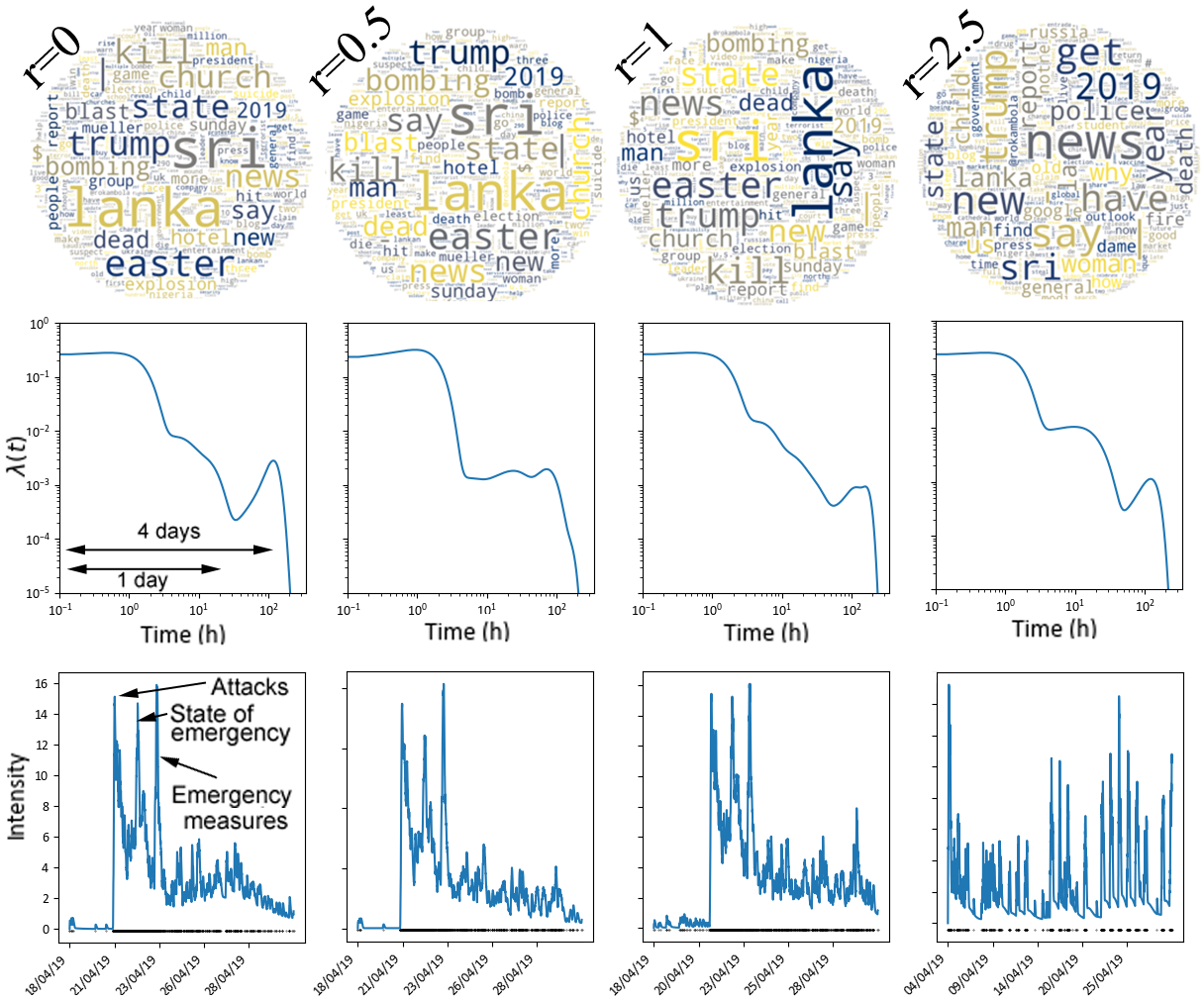}
    \caption{\textbf{Un cluster tiré de PDHP appliqué à des titres de journaux : les bombardements au Sri Lanka} --- Nuages de mots de ce cluster, kernel $\lambda (t)$ (Eq.\ref{eq-HawkesClus}), et intensité des clusters de vocabulaire proche des bombardements du Sri Lanka en avril 2019 pour plusieurs $r$.
    }
    \label{fig-res-Srilanka}
\end{figure}

Sur des \textbf{données réelles}, notre méthode fournit des résultats satisfaisants. Elle parvient à extraire des séries de publications ayant trait aux événements majeurs d'avril 2019 : bombardements au Sri Lanka, arrestation de Julian Assange, image directe d'un trou noir, incendie de Notre-Dame, entre autres. En nous concentrant sur les bombardements au Sri Lanka Fig.\ref{fig-res-Srilanka}, notre modèle repère correctement les points forts de la crise les 21, 22 et 23 avril. Lorsque $r$ est grand, le modèle créée des clusters avec des documents suivant un rythme de publication quotidien, sans motif apparent dans le vocabulaire employé, comme attendu. Des images similaires pour d'autres sujets sont disponibles en ligne, avec le code et les données utilisées.

\section{Conclusion}
Nous avons développé le DHP dans une forme plus générale, PDHP, qui permet non seulement de prendre en compte la date de publications de documents dans leur clustering, mais également de choisir l'importance qui lui est donnée par rapport à son contenu textuel.
Nombre d'extensions sont possibles pour PDHP. Par exemple, en développer une version considérant des processus de Hawkes \textit{multivariés} permettrait d'étudier comment les thématiques s'influencent les unes les autres. Une application couplant PDHP avec un modèle de langue plus sophistiqué que Dirichlet-Multinomial serait également intéressant. Finalement, une méthode permettant d'inférer la valeur de $r$ en fonction d'une quantité à optimiser (réduire la variance textuelle des clusters, par exemple) permettrait de faire de PDHP, mais également de PDP, des \textit{a priori} objectifs décrivant au mieux le monde réel.

\bibliographystyle{rnti}
\bibliography{biblio_exemple}

\Fr

\end{document}